\documentclass[11pt]{article}

\usepackage{acl}

\usepackage{times}
\usepackage{tabularx}
\usepackage{latexsym}
\usepackage{amsmath}
\usepackage{booktabs}
\usepackage{multirow}
\usepackage{algorithm}
\usepackage{algpseudocode}
\usepackage[T1]{fontenc}

\usepackage[utf8]{inputenc}

\usepackage{microtype}

\usepackage{inconsolata}

\usepackage{graphicx}

\title{When to Use Extra Context: Evidence-Grounded Terminology Adaptation for Simultaneous Speech Translation}

\author{
  Zeyu Yang\textsuperscript{1}, 
  Satoshi Nakamura\textsuperscript{1}\\
  \textsuperscript{1}The Chinese University of Hong Kong, Shenzhen  \\
  \textbf{Correspondence:} \href{mailto:your.email@domain.com}{zeyuyang1@link.cuhk.edu.cn}, \href{mailto:author2@domain.com}{snakamura@cuhk.edu.cn}
}

\begin{document}
\maketitle
\begin{abstract}
Extra context is valuable for simultaneous speech translation of technical talks, but injecting the entire document context into every streaming segment is often too coarse. Through diagnostic experiments, we find that context gains mainly come from paper-specific terminology recovery rather than uniform semantic enhancement. We therefore propose EGTA, an Evidence-Grounded Terminology Adaptation framework that builds a document terminology memory, selects compact candidate terms conditioned on the current streaming state, and adapts ASR/speech-side and decoder-side decision spaces using only the selected terms. EGTA can be instantiated in cascaded, end-to-end, and generation-only SimulST settings without full-model fine-tuning. We evaluate EGTA on an ACL technical-talk SimulST evaluation suite consisting of MCIF-dev and ACL60/60-dev. On MCIF-dev, EGTA-RG improves BLEU by +1.05/+0.59, XCOMET-XL by +0.019/+0.006, named-entity recall by +79\%/+73\% relative, and acronym recall by +0.099/+0.171 on En$\rightarrow$Zh and En$\rightarrow$De. Across MCIF-dev latency settings, EGTA consistently improves XCOMET-XL, named-entity recall, and acronym recall. External validation on ACL60/60-dev further shows consistent terminology-recall gains without additional fine-tuning. Shuffled-memory controls and activation audits provide evidence that the improvements are tied to paper-specific evidence alignment rather than generic context prompting.
\end{abstract}

\section{Introduction}

Simultaneous speech translation (SimulST) requires a system to translate speech before the full utterance or document is observed. This setting is particularly challenging for technical talks, where the speaker frequently mentions paper-specific terminology such as acronyms, model names, datasets, method names, and named entities. Unlike ordinary lexical choices, these terms are often rare, newly coined, or domain-specific. Missing or mistranslating them can substantially reduce the usefulness of the translated stream, even when the surrounding sentence is fluent.

A natural way to address this problem is to provide the model with extra document context, such as the talk title, abstract, or extracted entities. However, how to use such context in SimulST remains underexplored. A straightforward solution is to inject the entire context into every streaming segment, either through prompting, retrieval augmentation, or hidden-state adaptation. Our diagnostic experiments show that this uniform strategy is often too coarse: extra context can alter surface style or domain register without consistently improving the terminology fidelity that actually benefits from context. In streaming translation, many segments require no document-specific term, while only segments that mention paper-specific terms require precise terminology preservation.

This observation motivates a shift from general semantic context injection to terminology-oriented, evidence-conditioned adaptation. Instead of asking whether the model should always use document context, we ask: \emph{when should extra context be used, and where should it be applied?} We find that the most reliable gains from extra context are concentrated on paper-specific terminology rather than uniform semantic enhancement. Therefore, context adaptation should be selective, conditioned on the current streaming state, and applied to the decision spaces where terminology errors arise.

We propose \textbf{EGTA}, an \textbf{Evidence-Grounded Terminology Adaptation} framework for extra-context SimulST. EGTA first constructs a document terminology memory from paper-level context, including acronyms, model names, dataset names, method names, and other paper-specific entities. For each streaming segment, EGTA selects a compact candidate set from this memory using the current streaming state, such as the audio-derived hypothesis, transcript context, or partial translation state. It then applies the selected candidates to ASR/speech-side and decoder-side decision spaces. ASR/speech-side adaptation helps the system hear or attend to relevant terminology, while decoder-side adaptation biases decoding toward selected terminology tokens. We use ``evidence-grounded'' as a behavioral claim: the model should realize adapted terms only when they are supported by speech-conditioned evidence, which we later validate through activation audits and shuffled-memory controls.

EGTA is architecture-agnostic. The same selected term set can be exposed through ASR hotwords in cascaded systems, compact terminology context in end-to-end speech-language models, or decoder-side logit bias when only the generation interface is editable. This makes EGTA a lightweight inference-time framework rather than a full-model fine-tuning method.

We evaluate EGTA on MCIF-dev and ACL60/60-dev for En$\rightarrow$Zh and En$\rightarrow$De. With a fixed Qwen3-Omni-based end-to-end backbone, EGTA-RG with $B=2$ improves BLEU, XCOMET-XL, named-entity recall, and acronym recall at the MCIF-dev anchor point, and ACL60/60-dev shows consistent terminology-recall gains without additional fine-tuning. Latency-frontier results, shuffled-memory controls, and activation audits further indicate that the gains come from correct paper-specific evidence alignment rather than generic context prompting.

Our contributions are summarized as follows:
\begin{itemize}
    \item We present a diagnostic analysis showing that the main benefit of extra context in technical-talk SimulST comes from paper-specific terminology recovery, while uniform context injection can be unreliable.
    \item We propose EGTA, an evidence-grounded terminology adaptation framework that restricts inference-time adaptation to compact candidate terms selected from document context and the current streaming state.
    \item We analyze EGTA across end-to-end, cascaded, and generation-only backbones, showing where terminology adaptation provides the most headroom without full-model fine-tuning.
    \item Experiments on an ACL technical-talk evaluation suite show consistent improvements in XCOMET-XL, named-entity recall, and acronym recall, with external ACL60/60-dev validation and grounding audits supporting paper-specific evidence alignment.
\end{itemize}
\section{Related Work}
\label{sec:related-work}

\subsection{Simultaneous Speech Translation}

Simultaneous speech translation (SimulST) requires balancing quality and latency while the source speech is still unfolding. Prior work has studied read/write policies such as wait-$k$, monotonic attention, and information-transport policies, as well as evaluation toolkits and latency metrics such as SimulEval and length-adaptive average lagging \citep{ma-etal-2019-stacl, arivazhagan-etal-2019-monotonic, ma-etal-2020-simuleval, zhang-feng-2022-information, papi-etal-2022-generation}. These methods primarily improve segmentation, policy learning, or latency-aware decoding. Our work is complementary: we study how to use extra document context when only a subset of streaming segments requires paper-specific terminology.

\subsection{Context-Aware and Terminology-Aware Translation}

Context-aware translation has been widely studied in document-level MT and ST, including prior-sentence context, document-level context, and analyses of whether models actually use contextual signals \citep{voita-etal-2018-context, maruf-etal-2019-selective, maruf-etal-2021-survey, zhang-etal-2021-beyond, dabre-etal-2021-studying, fernandes-etal-2021-measuring}. Terminology adaptation has also been studied through lexical constraints, dictionary injection, terminology-aware training, and vocabulary biasing \citep{hokamp-liu-2017-lexically, post-vilar-2018-fast, dinu-etal-2019-training}. These works show that broader context or external lexical knowledge can improve translation, especially in domain-specific settings.

SimulST creates a different challenge: context must be useful before the full utterance or document is observed. Uniformly injecting all document terms into every segment may expose the model to irrelevant context and encourage hallucinated terminology. EGTA therefore differs from general document-context methods by treating context as a sparse terminology memory and activating it only when the current speech stream provides supporting evidence.

\subsection{Speech-Language Models and Contextual Biasing}

End-to-end speech translation and speech-language models, including Translatotron, SpeechT5, and SeamlessM4T, directly map speech to target text or unify speech and text modeling \citep{jia-etal-2019-direct, ao-etal-2022-speecht5, barrault2023seamlessm4t}. In speech recognition, contextual biasing and hotwording are commonly used to improve recognition of rare names, technical terms, and personalized vocabulary \citep{pundak-etal-2018-deep}. EGTA connects these ideas by decomposing terminology adaptation into ASR/speech-side and decoder-side decision spaces, instead of treating document context as a uniform signal applied to every segment.

In contrast to prior context-aware SimulST work that studies whether context helps at the sentence or document level, our focus is when context should be activated during streaming inference. EGTA operationalizes this question as terminology-memory construction, evidence-conditioned candidate selection, and architecture-specific ASR/speech-side and decoder-side adaptation.

\section{Method}
\label{sec:method}

\begin{figure*}[t]
    \centering
    \includegraphics[width=1\textwidth]{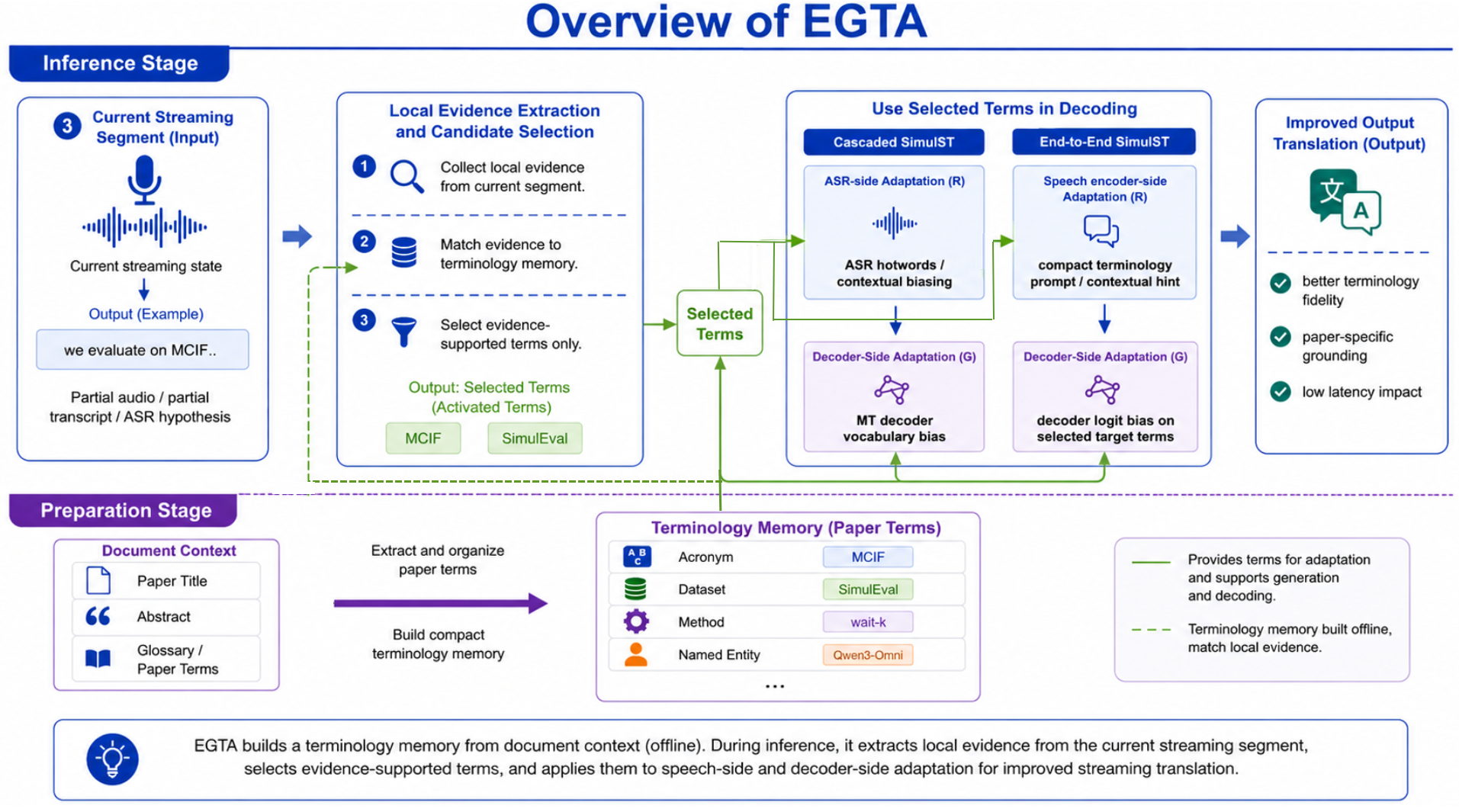}
    \caption{
    Overview of EGTA. The bottom panel shows the preparation stage, where document context is converted offline into a compact terminology memory. The top panel shows the inference stage: EGTA extracts local evidence from the current streaming segment, matches it to the terminology memory, selects evidence-supported terms, and uses them as inference-time conditioning signals for ASR-side, speech-encoder-side, and decoder-side decisions. Local evidence refers to a cue from the current segment, such as a partial transcript or ASR hypothesis that matches a term in the terminology memory.
    }
    \label{fig:egta_overview}
\end{figure*}

Figure~\ref{fig:egta_overview} shows the two-stage EGTA workflow. This section specifies the terminology memory, the evidence-conditioned selection algorithm, and the concrete R/G adaptation interfaces used in our systems.

\subsection{Problem Setup}
\label{sec:problem-setup}

Given an input speech stream $X$, a SimulST system incrementally receives audio chunks and emits target-language text before the full utterance or talk is observed. For a streaming segment $i$, the system observes partial speech input $X_{\leq i}$ and produces target output $Y_i$. In technical talks, we additionally assume access to document-level context $D$, such as the paper title, abstract, glossary, or automatically extracted metadata.

The goal is to use $D$ to improve terminology fidelity without introducing irrelevant terms into segments where the context is not needed. A naive approach is to concatenate the whole document context to every streaming input. However, this treats all segments as equally context-dependent. EGTA instead treats document context as a sparse terminology source and restricts adaptation to a segment-specific candidate subset.

\subsection{Terminology Memory}
\label{sec:term-memory}

EGTA converts document context $D$ into a terminology memory $T_D=\{t_1,\ldots,t_m\}$ containing paper-specific terms that are difficult for speech translation, including acronyms, model and dataset names, method names, organization names, and technical phrases. The memory is not taken from an IWSLT-provided gold glossary. We construct it automatically from paper-level context, including titles, abstracts, and available metadata, using a prompt-based Qwen3-30B-Instruct extractor served in FP8. The extractor proposes candidate named entities and technical terms, after which we normalize terms by case, punctuation, whitespace, and hyphen variants. We keep acronym-like strings, alphanumeric or hyphenated technical names, CamelCase terms, and multi-word technical phrases, while filtering generic function words and non-technical metadata. Candidate named entities are therefore treated as a noisy terminology list rather than a gold inventory; EGTA uses local speech evidence to select from this list instead of trusting every extracted term. Each term stores surface forms and tokenizer-level variants for both ASR/speech-side conditioning and decoder-side token biasing. Appendix~\ref{app:term-stats} reports memory-size and term-type statistics.

\subsection{Evidence-Conditioned Candidate Selection}
\label{sec:term-activation}

For each streaming segment, EGTA selects a compact subset of terms from the document memory. Let $E_i$ denote the local evidence for segment $i$, including the current speech segment, partial transcript or ASR hypothesis, nearby history, and partial target-side state. EGTA forms
\begin{equation}
T_i = A(T_D, E_i),
\end{equation}
where $A(\cdot)$ is the deterministic inference-time procedure in Algorithm~\ref{alg:egta}. The selector normalizes $E_i$ with the same rules as $T_D$, matches surface forms, acronyms, and tokenizer-level variants, and ranks matches by exactness, specificity, and recency under a fixed per-segment terminology budget. If the current partial hypothesis contains ``we evaluate on MCIF'', for example, the mention of ``MCIF'' is an evidence event that activates the memory term \textit{MCIF}. Detailed matching rules and prompt formatting are given in Appendix~\ref{app:implementation}.

\paragraph{Running example.}
Consider a segment whose partial ASR hypothesis contains ``we evaluate on MCIF'' while the document memory contains terms such as \textit{MCIF}, \textit{SimulEval}, and \textit{Qwen3-Omni}. The selector treats the observed mention of \textit{MCIF} as local evidence, ranks it above unmatched memory terms, and exposes only the selected set to the downstream interfaces. EGTA-R inserts the selected terms into the compact terminology context or ASR hotword list, while EGTA-G biases only the corresponding decoder tokens. Unmatched memory terms remain available for later segments but are not activated for the current one.
\begin{algorithm}[t]
\footnotesize
\caption{EGTA inference for streaming segment $i$}
\label{alg:egta}
\begin{algorithmic}[1]
\Require document context $D$, streaming state $E_i$, terminology budget $K$, bias strength $B$
\State Build or load terminology memory $T_D$ from $D$
\State Normalize the ASR hypothesis, partial transcript, and recent history in $E_i$
\State Match terms in $T_D$ by surface form, acronym, or tokenizer-level variant
\State Rank matches by exact match, acronym match, specificity, and recency
\State Select top-$K$ evidence-supported terms $T_i$
\State Apply $R(T_i)$ as ASR hotwords or compact terminology context
\State Apply $G(T_i)$ as decoder-side vocabulary bias with strength $B$
\State Decode the next streaming output prefix
\end{algorithmic}
\end{algorithm}

The selected terms define what is exposed to ASR/speech-side conditioning and decoder-side biasing, but realization is still determined by the speech-conditioned decoder. We therefore use ``evidence-grounded'' as a behavioral claim rather than a hard symbolic guarantee, and validate it through activation audits, evidence-strength buckets, and shuffled-memory controls in Section~\ref{sec:grounding-validation}.

\subsection{ASR/Speech-Side and Decoder-Side Adaptation}
\label{sec:rg-adaptation}

EGTA applies selected terms through two interfaces: $R(T_i)$ for ASR/speech-side conditioning and $G(T_i)$ for decoder-side biasing. EGTA-RG combines both.

\paragraph{ASR/speech-side adaptation.}
ASR/speech-side adaptation helps the system hear or attend to relevant terminology before target decoding. In cascaded systems, $R(T_i)$ is implemented through ASR hotwords or contextual biasing. In end-to-end systems, $R(T_i)$ is implemented as inference-time terminology conditioning for the speech-language model rather than as a parameter update. It exposes only selected terms, e.g., \textit{Relevant paper terms: MCIF; SimulEval}, instead of the full document context.

\paragraph{Decoder-side adaptation.}
Decoder-side adaptation preserves selected terminology during target decoding. For each selected term $t \in T_i$, EGTA identifies target-side token variants and applies a vocabulary-level bias. Let $z_t(v)$ be the original logit for vocabulary token $v$ at decoding step $t$. EGTA-G modifies it as
\begin{equation}
z'_t(v) = z_t(v) + B \cdot \mathbf{1}[v \in V(T_i)],
\end{equation}
where $V(T_i)$ is the set of vocabulary tokens associated with the selected terms and $B$ is the bias strength. Target-side variants are constructed from the terminology memory and tokenizer outputs, not from test references. This biasing rule is deliberately simple: our goal is not to introduce a new constrained-decoding algorithm, but to test whether document terms should be activated selectively under streaming evidence. We therefore compare EGTA-G with stronger alternatives, including uniform full-memory context and Global-G, which applies decoder-side bias to all memory terms without evidence-conditioned selection. As shown in Section~\ref{sec:additional-analyses} and Appendix~\ref{app:uniform-global}, stronger all-term bias increases over-biasing risk and can degrade BLEU and latency, while selected-term bias gives a better recall--quality--latency trade-off. Unlike global terminology biasing, EGTA-G biases only evidence-selected terms, limiting over-biasing on unrelated segments.

\paragraph{Full EGTA-RG.}
When both interfaces are available, EGTA decodes with
\begin{equation}
Y_i = M_{\theta}(X_{\leq i}; R(T_i), G(T_i)).
\end{equation}
Here $R(T_i)$ improves access to relevant terminology in the speech-understanding stage, while $G(T_i)$ preserves selected terms through token-level biasing.

\subsection{Architecture Instantiations}
\label{sec:architecture-instantiations}

Table~\ref{tab:instantiation} summarizes the concrete interfaces used in each backbone. All instantiations share the same selected term set $T_i$; they differ only in how $R(T_i)$ and $G(T_i)$ are exposed by the model.

\begin{table}[t]
\centering
\footnotesize
\setlength{\tabcolsep}{3pt}
\resizebox{\columnwidth}{!}{%
\begin{tabular}{@{}lll@{}}
\toprule
Backbone & $R(T_i)$ interface & $G(T_i)$ interface \\
\midrule
Qwen3-Omni E2E & compact term context & decoder logit bias \\
Cascade ASR--MT & ASR hotwords & MT vocabulary bias \\
SeamlessM4T & -- & decoder-side EGTA-G \\
\bottomrule
\end{tabular}%
}
\caption{Concrete EGTA instantiations across backbones.}
\label{tab:instantiation}
\end{table}

The Qwen3-Omni end-to-end setting is our main system. The cascade setting diagnoses how much headroom remains when ASR-side hotwording is already available, and SeamlessM4T provides a generation-only transfer check when the $R(T_i)$ interface is unavailable.

\section{Experimental Setup}
\label{sec:setup}

\subsection{Datasets}

We evaluate EGTA on an ACL technical-talk SimulST evaluation suite consisting of MCIF-dev and ACL60/60-dev. Both subsets are built from technical conference talks with paper-level context, making them suitable for evaluating terminology adaptation under streaming speech evidence.

MCIF-dev is derived from MCIF (Multimodal Crosslingual Instruction Following), a multilingual benchmark based on scientific talks \citep{papi2025mcif}. We pair its speech and references with paper-level metadata, including titles, abstracts, and extracted terminology, and use the resulting 21-talk, 919-segment technical-talk subset as the main benchmark for full latency, ablation, and grounding analyses.

ACL60/60-dev is used as an external technical-talk benchmark. It contains 5 ACL talks and 468 gold-segmented segments with official En$\rightarrow$Zh and En$\rightarrow$De references. It also provides tagged terminology for both target languages, allowing terminology-sensitive evaluation beyond MCIF-dev. We use the same terminology extraction, normalization, candidate selection, and EGTA-RG B=2 configuration without additional fine-tuning.

\subsection{Systems and Configurations}
For end-to-end SimulST, we use a fine-tuned Qwen3-Omni-30B-A3B audio-text LLM as the backbone. Following our IWSLT system setup, the audio encoder is kept frozen, the non-text generation modules are disabled, and only language-side LoRA adapters are used to teach streaming read/write behavior. At inference time, the backbone is served with vLLM using greedy decoding and a bounded dialogue history. The backbone checkpoint, chunk schedule, decoding parameters, and evaluation talks are fixed across variants; variants differ only in the inference-time terminology interface. The baseline receives the streaming speech input and task instruction without document terminology adaptation. We additionally include a uniform-context baseline, which injects the full per-paper terminology memory into every streaming segment without evidence-conditioned selection or decoder-side biasing. This baseline tests whether simply exposing document terminology to every segment is sufficient.

EGTA-R adds a compact terminology block selected from the document memory; EGTA-G applies decoder-side vocabulary bias to terminology tokens; and EGTA-RG combines both channels. Unless otherwise specified, our recommended end-to-end configuration is EGTA-RG with bias strength $B=2$.

We also evaluate EGTA in a cascaded ASR--MT setting for architecture analysis. We use Qwen3-ASR-1.7B as the ASR front end and pass the selected terms as ASR hotwords/contextual-biasing terms for EGTA-R; EGTA-G is implemented through MT decoder vocabulary bias. Finally, we evaluate SeamlessM4T-v2-large as a generation-only backbone validation. Since this backbone does not expose the same editable terminology-prompting interface used for EGTA-R, we instantiate only EGTA-G. To test whether gains are paper-specific, we additionally use a shuffled-memory control that pairs each segment with terminology memory from a mismatched paper.

\subsection{Metrics and Diagnostics}

We report BLEU, chrF, XCOMET-XL, NER$_{\text{clean}}$ recall, acronym recall, and LongLAAL$_{\mathrm{CU}}$ latency. BLEU and chrF measure surface overlap, XCOMET-XL provides a neural quality estimate, and lower LongLAAL$_{\mathrm{CU}}$ indicates lower latency. NER$_{\text{clean}}$ recall is computed over cleaned paper-specific entities after normalizing case, punctuation, whitespace, hyphenation, and tokenizer artifacts; acronym recall is computed over uppercase and alphanumeric paper terms. We use terminology recall as the primary metric because omission or mistranslation of paper-specific terms is the main failure mode in technical-talk SimulST. Since surface-form precision depends on whether a reference preserves the same term realization, precision/F1-style matching is treated as a conservative diagnostic, while possible over-generation is analyzed through unsupported-activation audits, shuffled-memory controls, and Global-G stress tests.

For the main end-to-end experiments, we evaluate five latency operating points; the anchor point is used for detailed comparison, ablation, and shuffled-memory control. In addition to aggregate metrics, we conduct activation audits, evidence-strength bucket analysis, and shuffled-memory controls to validate whether realized terminology usage is supported by speech-conditioned evidence rather than generic context exposure.

\section{Results and Analysis}
\label{sec:results}

\subsection{Main End-to-End Results}
\label{sec:main-e2e-results}

Table~\ref{tab:e2e_anchor} reports anchor-point gains of EGTA-RG over the baseline on the ACL technical-talk evaluation suite. The main pattern is consistent across both datasets and both language directions: EGTA-RG improves terminology-sensitive metrics, while maintaining or improving overall translation quality.

\begin{table}[t]
\centering
\footnotesize
\setlength{\tabcolsep}{3pt}
\resizebox{\columnwidth}{!}{%
\begin{tabular}{@{}llccccc@{}}
\toprule
Set & Lang. & $\Delta$BLEU & $\Delta$XCOMET & $\Delta$NER & $\Delta$Acro & $\Delta$LAAL \\
\midrule
MCIF-dev & Zh & +1.05 & +.019 & +.212 & +.099 & -46 \\
MCIF-dev & De & +0.59 & +.006 & +.266 & +.171 & -44 \\
ACL60/60-dev & Zh & +0.94 & +.022 & +.113 & +.119 & -- \\
ACL60/60-dev & De & +0.56 & +.005 & +.214 & +.100 & -- \\
\bottomrule
\end{tabular}%
}
\caption{
Anchor-point gains of EGTA-RG B=2 over the baseline on the ACL technical-talk evaluation suite. NER denotes NER$_{\text{clean}}$ recall. LAAL is reported for MCIF-dev, where the streaming-yaml protocol is matched across systems. Full absolute scores are reported in Appendix~\ref{app:full-anchor}.
}
\label{tab:e2e_anchor}
\end{table}

On MCIF-dev, EGTA-RG improves all reported anchor-point metrics on both En$\rightarrow$Zh and En$\rightarrow$De. The largest gains appear on terminology fidelity: NER$_{\text{clean}}$ recall improves by +0.212/+0.266 and acronym recall by +0.099/+0.171, while XCOMET-XL also increases by +0.019/+0.006. Latency is slightly reduced on both directions, with LongLAAL$_{\mathrm{CU}}$ decreasing by 46 ms and 44 ms.

ACL60/60-dev provides an external validation of the same trend without additional fine-tuning. EGTA-RG improves NER$_{\text{clean}}$ recall by +0.113/+0.214 and acronym recall by +0.119/+0.100 on En$\rightarrow$Zh and En$\rightarrow$De. On En$\rightarrow$Zh, it also improves BLEU by +0.94 and XCOMET-XL by +0.022 over the baseline and outperforms uniform context on overall quality.

We further conduct paired bootstrap resampling with 5,000 samples at both segment and talk levels on MCIF-dev. The terminology-fidelity gains are statistically robust on both language pairs: NER$_{\text{clean}}$ recall and acronym recall improve significantly on En$\rightarrow$Zh and En$\rightarrow$De ($p<0.001$). Appendix~\ref{app:acl6060} reports ACL60/60-dev significance details, and Appendix~\ref{app:precision-f1} reports reference-supported terminology matching. Overall, these results support our central hypothesis: extra document context is most useful when converted into selective terminology adaptation rather than injected uniformly as general semantic context.

\subsection{End-to-End Latency--Quality Frontier}
\label{sec:e2e-frontier}

\begin{figure*}[t]
    \centering
    \includegraphics[width=0.92\textwidth]{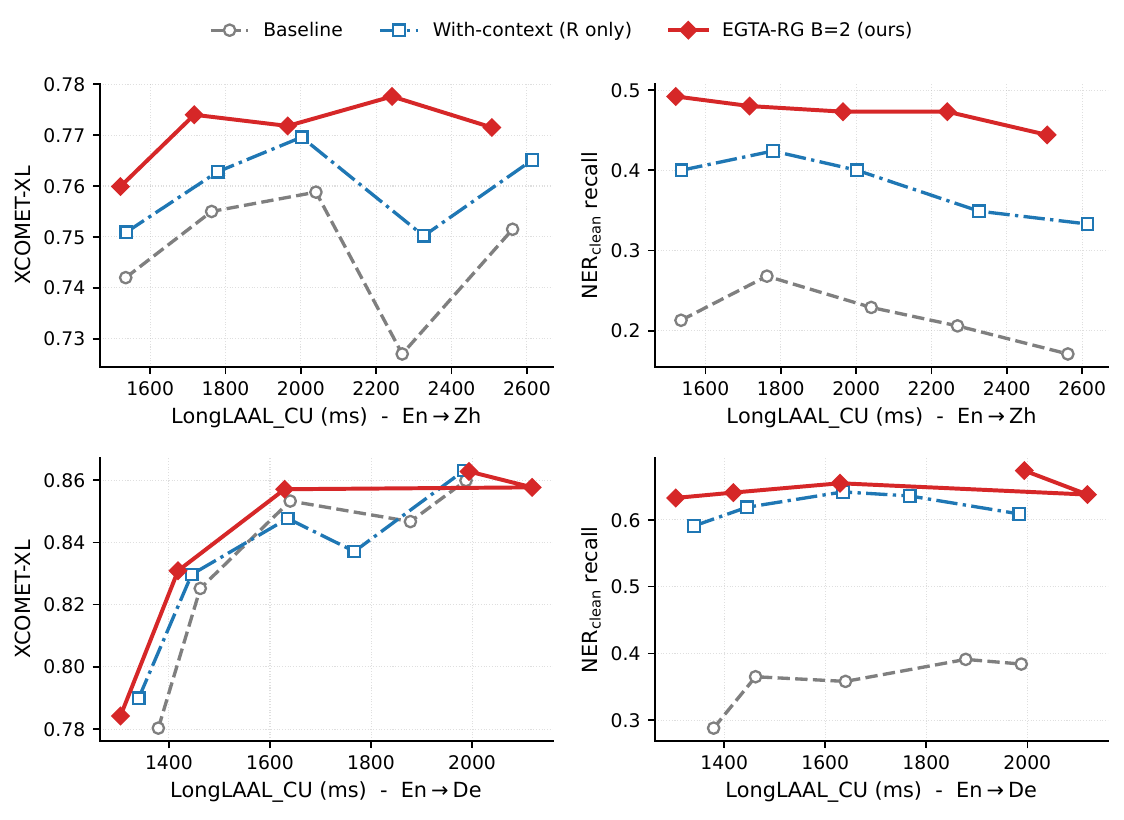}
    \caption{
    End-to-end latency--quality frontier on MCIF-dev. Each marker corresponds to one latency operating point. The top row shows En$\rightarrow$Zh and the bottom row shows En$\rightarrow$De; the left column reports XCOMET-XL and the right column reports NER$_{\text{clean}}$ recall. EGTA-RG B=2 consistently improves XCOMET-XL and terminology recall over the baseline, and generally outperforms the uniform-context variant on terminology recall.
    }
    \label{fig:e2e_frontier}
\end{figure*}

On MCIF-dev, we further evaluate EGTA-RG across five latency operating points. Figure~\ref{fig:e2e_frontier} shows the end-to-end latency--quality frontier on XCOMET-XL and NER$_{\text{clean}}$ recall. EGTA-RG consistently outperforms the baseline on both metrics across operating points, and is generally stronger than the uniform-context variant, especially on terminology recall. This indicates that simply exposing the full terminology memory to every segment recovers part of the benefit, but does not provide the same terminology--latency trade-off as evidence-conditioned R+G composition. We report only end-to-end systems in this figure; cascaded systems are analyzed separately in Section~\ref{sec:additional-analyses} to avoid mixing different latency sources and terminology-recall protocols.

Across the five operating points, EGTA-RG improves the En$\rightarrow$Zh baseline by +1.31 BLEU, +2.30 chrF, +0.0241 XCOMET-XL, +0.255 NER$_{\text{clean}}$ recall, and +0.117 acronym recall on average, while reducing LongLAAL$_{\mathrm{CU}}$ by 43 ms. On En$\rightarrow$De, EGTA-RG improves BLEU by +0.81, XCOMET-XL by +0.0055, NER$_{\text{clean}}$ recall by +0.277, and acronym recall by +0.154 on average. XCOMET-XL, NER$_{\text{clean}}$, and acronym recall improve at all five operating points, and the positive average BLEU gain suggests that the terminology improvements do not come at the cost of overall translation quality.

\begin{table}[!t]
\centering
\footnotesize
\setlength{\tabcolsep}{3pt}
\renewcommand{\arraystretch}{1.05}
\begin{tabular}{@{}p{0.25\columnwidth}p{0.43\columnwidth}p{0.24\columnwidth}@{}}
\toprule
Analysis & Main observation & Takeaway \\
\midrule
Uniform context & Partial gains; below EGTA-RG & Selection helps \\
Global-G $B=5$ & Zh BLEU $-2.1$; latency collapse & Over-bias stress \\
Cascaded SimulST & Zh: +0.25 BLEU, +1.46 NER & Less headroom \\
SeamlessM4T & De NER: 36.7$\rightarrow$51.7 & Decoder transfer \\
\bottomrule
\end{tabular}
\caption{
Auxiliary analyses testing always-on context, over-biasing, and architectural transfer. Full metrics are provided in Appendices~\ref{app:uniform-global} and~\ref{app:seamless-cost}.
}
\label{tab:additional_analyses}
\end{table}

\subsection{Additional Analyses}
\label{sec:additional-analyses}

Table~\ref{tab:additional_analyses} summarizes four auxiliary analyses.
\textit{Uniform context} exposes the full paper terminology memory to every segment.
\textit{Global-G} applies decoder-side vocabulary bias to all memory terms without evidence-conditioned selection and serves as an over-bias stress test rather than a matched operating-point baseline.

Together, these checks show that EGTA is not merely stronger context injection: always-on context or biasing is less stable, while architecture checks show that gains depend on remaining terminology headroom. Shuffled-memory controls further suggest that the improvements are paper-specific rather than generic biasing effects.

\subsection{Grounding Validation}
\label{sec:grounding-validation}

Table~\ref{tab:grounding} summarizes three grounding diagnostics. EGTA usually activates terms supported by the current speech stream, and its largest gains appear in segments with explicit evidence.

\begin{table}[!t]
\centering
\footnotesize
\setlength{\tabcolsep}{4pt}
\begin{tabular}{@{}lcc@{}}
\toprule
Diagnostic & En$\rightarrow$Zh & En$\rightarrow$De \\
\midrule
Local/ref audit support & 86.4\% & 86.1\% \\
Same-paper audit support & 96.0\% & 93.9\% \\
Unsupported activation & 4.0\% & 6.1\% \\
\midrule
High evid. $\Delta$chrF / $\Delta$NER & +3.7 / +27.7 & +1.8 / +16.2 \\
Medium evid. $\Delta$chrF / $\Delta$NER & +6.9 / +59.6 & +3.2 / +58.2 \\
Low/none $\Delta$chrF / Unsup. act. & +0.3 / 39.5\% & +0.4 / 27.3\% \\
\bottomrule
\end{tabular}
\caption{
Grounding validation. The first block audits local/reference support, same-paper support, and unsupported activation; references are used only for post-hoc validation. The second block reports gains by evidence strength. $\Delta$NER is in percentage points; Unsup. act. denotes unsupported activation rate.
}
\label{tab:grounding}
\end{table}

\paragraph{Activation support.}
The first block audits whether activated terms have post-hoc evidence. Local/reference audit support counts an activation as supported when the normalized term appears in local streaming evidence, such as the ASR hypothesis or nearby transcript, or in the segment-level reference. The reference is used only for validation, not for candidate selection or decoding. Same-paper audit support additionally allows the term to appear elsewhere in the same talk evidence. Unsupported activations account for only 4.0\% / 6.1\% of activations.

\paragraph{Evidence-strength buckets.}
We bucket segments before scoring outputs: high evidence has an exact normalized match in the local hypothesis or nearby transcript; medium evidence has an acronym, partial phrase, or tokenizer-level variant match; low/none has no such evidence. These buckets are computed from source-side evidence and the terminology memory, not from system outputs. High- and medium-evidence segments show the largest NER gains, while low/none-evidence segments show only small chrF changes (+0.3/+0.4), suggesting limited over-biasing on unrelated segments.

\paragraph{Shuffled-memory control.}
The shuffled-memory control verifies that the gains require the correct paper memory. On En$\rightarrow$De, the anchor NER gain drops from +0.266 with correct memory to +0.012 with shuffled memory, and the medium-evidence NER gain drops from +58.2 to +7.5 points. Thus, grounding is a behavioral property that emerges when speech-conditioned decoding is paired with the correct paper-specific terminology memory.

\section{Conclusion}
\label{sec:conclusion}

We presented EGTA, a lightweight inference-time framework for evidence-conditioned terminology adaptation in technical-talk SimulST. Across MCIF-dev and ACL60/60-dev, EGTA improves terminology recall while maintaining overall translation quality. Latency-frontier results, shuffled-memory controls, and activation audits suggest that extra context is most useful when exposed as selected terminology evidence rather than injected uniformly into every segment.

\section*{Limitations}

This work focuses on technical-talk SimulST, where document-level paper context is available and terminology fidelity is important; open-domain, informal, and low-context settings remain to be studied. EGTA also depends on terminology-memory quality: missing or irrelevant extracted terms can weaken both ASR/speech-side and decoder-side adaptation.

The grounding criterion is behavioral and audit-based rather than enforced by a separately trained symbolic evidence detector. The current selector uses normalized surface matching, so severe ASR errors or phonetic variants may prevent term activation; future work should add phonetic or fuzzy matching.

Our evaluation emphasizes automatic terminology-recall metrics, complemented by activation audits, shuffled-memory controls, over-bias diagnostics, and precision/F1-style surface matching. We do not include full human evaluation of terminology correctness, perceived usefulness, or harmful wrong insertions. Finally, both MCIF-dev and ACL60/60-dev are technical conference-talk subsets, and our SeamlessM4T validation covers only decoder-side EGTA-G because the backbone does not expose the editable EGTA-R interface.

\section*{Ethical Considerations}

EGTA aims to improve terminology fidelity in technical-talk SimulST by preserving names, acronyms, datasets, and method terms. However, terminology adaptation can amplify errors if document context is mismatched, outdated, or maliciously constructed. The system may still produce mistranslations, omissions, or hallucinated terms, especially in low-evidence segments or unreliable speech conditions, and should not be used as the sole mechanism in high-stakes settings without human oversight. Our experiments use technical-talk data and paper-level context and do not involve personal profiling beyond the context required for translation.
\bibliography{custom}

\appendix

\section{Additional Experimental Analyses}
\label{sec:appendix}

\subsection{Full Anchor-Point Results}
\label{app:full-anchor}

Table~\ref{tab:app_full_anchor} reports the absolute anchor-point scores corresponding to the gain summary in Table~\ref{tab:e2e_anchor}. The main text reports compact gains for readability, while this table provides the underlying baseline and EGTA-RG scores.

\begin{table*}[t]
\centering
\footnotesize
\setlength{\tabcolsep}{4pt}
\begin{tabular}{@{}lllcccccc@{}}
\toprule
Set & Lang. & System & BLEU & chrF & XCOMET & NER & Acro & LAAL \\
\midrule
MCIF-dev & Zh & Baseline & 43.31 & 36.67 & 0.7550 & 0.268 & 0.680 & 1763 \\
MCIF-dev & Zh & EGTA-RG B=2 & 44.36 & 38.56 & 0.7740 & 0.480 & 0.779 & 1717 \\
MCIF-dev & De & Baseline & 26.58 & 60.87 & 0.8252 & 0.365 & 0.618 & 1462 \\
MCIF-dev & De & EGTA-RG B=2 & 27.16 & 61.70 & 0.8309 & 0.631 & 0.789 & 1418 \\
\midrule
ACL60/60-dev & Zh & Baseline & 42.35 & 35.91 & 0.7337 & 0.415 & 0.616 & -- \\
ACL60/60-dev & Zh & EGTA-RG B=2 & 43.29 & 37.28 & 0.7554 & 0.528 & 0.734 & -- \\
ACL60/60-dev & De & Baseline & 28.29 & 61.00 & 0.8214 & 0.396 & 0.646 & -- \\
ACL60/60-dev & De & EGTA-RG B=2 & 28.85 & 61.17 & 0.8263 & 0.610 & 0.746 & -- \\
\bottomrule
\end{tabular}
\caption{
Full anchor-point results corresponding to Table~\ref{tab:e2e_anchor}. NER denotes NER$_{\text{clean}}$ recall. LAAL denotes LongLAAL$_{\mathrm{CU}}$ in milliseconds and is reported for MCIF-dev, where the streaming-yaml protocol is matched across systems.
}
\label{tab:app_full_anchor}
\end{table*}

\subsection{Terminology Memory Statistics}
\label{app:term-stats}

Table~\ref{tab:app_term_stats} summarizes the scale of terminology memory construction and per-segment terminology exposure. These statistics characterize the preparation stage and make explicit the number of extracted and selected terms used by EGTA. The same extraction, normalization, and ranking rules are used for all systems and operating points.

\begin{table}[t]
\centering
\footnotesize
\setlength{\tabcolsep}{3pt}
\resizebox{\columnwidth}{!}{%
\begin{tabular}{@{}lcc@{}}
\toprule
Statistic & MCIF-dev & ACL60/60-dev \\
\midrule
Talks & 21 & 5 \\
Segments & 919 & 468 \\
Selection budget $K$ & 10 & 10 \\
Bias strength $B$ & 2.0 & 2.0 \\
Raw NER entities / talk & 24.81 / 24 / 7 / 41 & 13.60 / 14 / 11 / 16 \\
Filtered $|M_D|$ / talk & 10.19 / 11 / 3 / 16 & 6.40 / 5 / 5 / 9 \\
Compact top-$K$ / talk & 8.71 / 10 / 3 / 10 & 5.80 / 5 / 4 / 8 \\
R-prompt terms / segment & 8.66 / 10 / 3 / 10 & 6.05 / 5 / 4 / 8 \\
G-bias candidates / segment & 10.09 / 11 / 3 / 16 & 6.59 / 5 / 5 / 9 \\
\bottomrule
\end{tabular}%
}
\caption{
Terminology-memory statistics. Values after the first four rows are reported as mean / median / min / max. $M_D$ denotes the filtered document terminology memory. The R-prompt exposes a compact top-$K$ term list, while G-bias candidates are vocabulary-bias candidates derived from the filtered memory.
}
\label{tab:app_term_stats}
\end{table}

Table~\ref{tab:app_term_types} reports the distribution of extracted term types after filtering. The categories correspond to the surface-form patterns retained by the terminology extraction rules.

\begin{table}[t]
\centering
\footnotesize
\setlength{\tabcolsep}{4pt}
\begin{tabular}{@{}lcc@{}}
\toprule
Term type & MCIF-dev & ACL60/60-dev \\
\midrule
Acronym-like & 32 (15.0\%) & 6 (18.8\%) \\
Alphanumeric / hyphenated & 6 (2.8\%) & 2 (6.2\%) \\
CamelCase / model or dataset & 4 (1.9\%) & 0 (0.0\%) \\
Multi-word technical & 31 (14.5\%) & 9 (28.1\%) \\
Named entity: person & 104 (48.6\%) & 15 (46.9\%) \\
Named entity: organization & 37 (17.3\%) & 0 (0.0\%) \\
\midrule
Total & 214 (100\%) & 32 (100\%) \\
\bottomrule
\end{tabular}
\caption{
Distribution of filtered terminology-memory types. The retained categories include acronym-like strings, alphanumeric or hyphenated technical names, CamelCase spans, model or dataset names, multi-word technical phrases, and named entities.
}
\label{tab:app_term_types}
\end{table}

\subsection{Implementation and Audit Details}
\label{app:implementation}

\paragraph{Candidate selection.}
For $A(T_D,E_i)$, we use deterministic normalized matching rather than fuzzy or phonetic matching. Memory terms and streaming hypotheses are normalized for Latin case, punctuation, whitespace, hyphen variants, tokenizer artifacts, digits, and acronym forms. A term is selected if its full surface form, acronym form, or tokenizer-level variant is matched in the current segment hypothesis or recent streaming history. For multiword terms, we match the full phrase, acronym, or content-bearing components; content-bearing components exclude generic function words and retain technical tokens such as acronyms, digits, CamelCase spans, hyphenated terms, and model or dataset names. When multiple candidates match, ties are resolved by preferring exact surface matches, then acronym matches, then longer and more specific technical terms. The same fixed per-segment terminology budget and ranking rule are used for all systems and operating points. The extracted terminology memory is produced by the Qwen3-30B-Instruct extractor plus deterministic filtering, and may contain extraction errors; EGTA is designed to select from this noisy memory using local speech evidence rather than to trust every extracted entity.

\paragraph{Target variants and prompt.}
For EGTA-G, $V(T_i)$ is built from the terminology memory and target tokenizer outputs, not from segment-level test references. Acronyms, model names, dataset names, and alphanumeric terms are kept as copyable forms across target languages. When a target alias is available in the memory, it is included as an additional variant; otherwise, the normalized source form and its tokenizer-level variants are used. Multi-token terms contribute their corresponding token ids to the bias set. For EGTA-R, selected terms are inserted into a compact terminology block attached to the task instruction and updated for each streaming segment, rather than exposing the full document memory.

\paragraph{Backbone and decoding details.}
The Qwen3-Omni backbone follows our IWSLT 2026 system setup. We use Qwen3-Omni-30B-A3B with the audio encoder frozen and LoRA adapters inserted only into the LLM/Thinker module; the vision and speech-synthesis modules are disabled. We train one LoRA checkpoint per target language on 220K retained chunk-level examples constructed from LibriSpeech, Common Voice 17.0, CoVoST2, and VoxPopuli. LoRA is applied to the \texttt{q\_proj}, \texttt{k\_proj}, \texttt{v\_proj}, and \texttt{o\_proj} modules with rank 16, alpha 32, and dropout 0.05. The peak learning rates are $1\times 10^{-4}$ for En$\rightarrow$Zh and $5\times 10^{-5}$ for En$\rightarrow$De, with global batch size 128 and 3 training epochs on 8 NVIDIA A100 GPUs. At inference time, the model is served with vLLM using greedy decoding; the streaming agent keeps the most recent 16 dialogue turns or 20 seconds of audio history. The latency operating points are controlled by the audio chunk size while all compared variants use the same backbone checkpoint and decoding settings.

\paragraph{Latency and grounding audit.}
LongLAAL$_{\mathrm{CU}}$ follows an LAAL-style computation over committed output units, using source chunk timestamps and emitted target prefixes. All systems compared at the same operating point use the same chunk schedule and segmentation policy, so latency differences reflect decoding and adaptation behavior rather than different input chunking. In grounding audits, an activation is counted as local/reference audit supported if its normalized surface form, acronym, or accepted tokenizer-level variant appears in either the local hypothesis available during streaming or the segment reference used for post-hoc validation. It is counted as same-paper audit supported if the same normalized term appears elsewhere in the same talk evidence. An unsupported activation is an activated paper term without local/reference support under these matching rules; near matches are counted only when they normalize to an accepted surface, acronym, or tokenizer-level variant.

\subsection{Statistical Robustness of Terminology Gains}
\label{app:significance}

We conduct paired bootstrap resampling with 5,000 samples at both segment and talk levels to assess whether the terminology-fidelity gains are robust. Table~\ref{tab:app_sig_terms} reports the results for the two primary terminology metrics. NER$_{\text{clean}}$ recall and acronym recall improve significantly on both language pairs, supporting our use of terminology recall as the primary evidence for EGTA's effect.

\begin{table}[t]
\centering
\footnotesize
\setlength{\tabcolsep}{4pt}
\begin{tabular}{@{}llccc@{}}
\toprule
Lang. & Metric & $\Delta$ & 95\% CI & $p$ \\
\midrule
En$\rightarrow$Zh & NER$_{\text{clean}}$ & +0.212 & [+0.140, +0.285] & $<0.001$ \\
En$\rightarrow$Zh & Acro & +0.099 & [+0.033, +0.168] & 0.004 \\
En$\rightarrow$De & NER$_{\text{clean}}$ & +0.266 & [+0.180, +0.351] & $<0.001$ \\
En$\rightarrow$De & Acro & +0.171 & [+0.103, +0.243] & $<0.001$ \\
\bottomrule
\end{tabular}
\caption{
Paired-bootstrap significance for terminology-fidelity metrics at the anchor operating point. $\Delta$ denotes EGTA-RG B=2 minus the baseline. Confidence intervals and $p$-values are computed with 5,000 segment-level bootstrap resamples.
}
\label{tab:app_sig_terms}
\end{table}

\subsection{ACL60/60-dev External Validation Details}
\label{app:acl6060}

ACL60/60-dev contains 5 ACL talks and 468 gold-segmented segments. We run the same anchor EGTA-RG B=2 configuration used for MCIF-dev without additional fine-tuning. Since this split does not provide the same gold streaming-yaml protocol as MCIF-dev, ACL60/60-dev latency is used only as an internal relative diagnostic; the main text focuses on text quality and terminology fidelity. Segment-level paired bootstrap shows significant En$\rightarrow$Zh gains in XCOMET-XL ($p=.006$), acronym recall ($p=.04$), and tagged-terminology recall ($p=.001$). On En$\rightarrow$De, NER$_{\text{clean}}$ and acronym gains are significant at both segment and talk levels.

\begin{table}[t]
\centering
\footnotesize
\setlength{\tabcolsep}{4pt}
\begin{tabular}{@{}llcccc@{}}
\toprule
Lang. & System & BLEU & chrF & XCOMET & GoldTag \\
\midrule
Zh & Baseline & 42.35 & 35.91 & 0.7337 & 0.709 \\
Zh & Uniform & 41.24 & 35.74 & 0.7289 & 0.728 \\
Zh & EGTA-RG B=2 & \textbf{43.29} & \textbf{37.28} & \textbf{0.7554} & \textbf{0.746} \\
\midrule
De & Baseline & 28.29 & 61.00 & 0.8214 & 0.656 \\
De & Uniform & 28.50 & 60.97 & 0.8123 & 0.631 \\
De & EGTA-RG B=2 & \textbf{28.85} & \textbf{61.17} & \textbf{0.8263} & 0.642 \\
\bottomrule
\end{tabular}
\caption{
ACL60/60-dev external validation. GoldTag denotes recall over official tagged terminology. ACL60/60-dev latency is used only as an internal diagnostic because its streaming-yaml protocol differs from MCIF-dev.
}
\label{tab:app_acl6060_full}
\end{table}

\subsection{Uniform Context and Global Bias Stress Tests}
\label{app:uniform-global}

We compare EGTA-RG against two stronger alternatives to test whether the gains can be explained by simply exposing more terminology or applying stronger decoder bias. The uniform-context baseline injects the full per-paper terminology memory into every streaming segment without evidence-conditioned selection or decoder-side biasing. Global-G applies decoder-side bias to all document terms for every segment without the ASR/speech-side prompt.

Table~\ref{tab:app_uniform_global} summarizes the En$\rightarrow$Zh anchor comparison. Uniform context recovers part of the gain, showing that document terminology is useful, but it does not match EGTA-RG on the joint recall--quality--latency trade-off. Global all-term biasing shows a clear over-bias failure mode when the bias is too strong: at $B=5$, BLEU drops below the baseline and latency collapses. In contrast, EGTA-RG with $B=2$ achieves the strongest overall trade-off among the tested configurations.

\begin{table}[t]
\centering
\scriptsize
\setlength{\tabcolsep}{2pt}
\resizebox{\columnwidth}{!}{%
\begin{tabular}{@{}lcccccc@{}}
\toprule
System & BLEU & chrF & XCOMET & NER & Acro & LAAL \\
\midrule
Baseline & 43.31 & 36.67 & 0.7550 & 0.268 & 0.680 & 1763 \\
Uniform & 44.08 & 38.35 & 0.7628 & 0.424 & 0.739 & 1779 \\
Global-G B=5 & 41.21 & 35.36 & 0.7364 & 0.398 & 0.685 & 1318 \\
EGTA-RG B=2 & \textbf{44.36} & \textbf{38.56} & \textbf{0.7740} & \textbf{0.480} & \textbf{0.779} & \textbf{1717} \\
\bottomrule
\end{tabular}%
}
\caption{
Uniform-context and global-bias stress test on En$\rightarrow$Zh. EGTA-RG B=2 provides the best recall--quality--latency trade-off; LAAL denotes LongLAAL$_{\mathrm{CU}}$ in milliseconds.
}
\label{tab:app_uniform_global}
\end{table}

We also conduct a bias-strength sweep on En$\rightarrow$Zh. G-only biasing improves terminology recall as $B$ increases, but aggressive biasing eventually harms overall quality. Once the ASR/speech-side prompt is active, increasing from EGTA-RG B=2 to B=3 yields diminishing returns: NER recall does not improve, while BLEU, XCOMET-XL, and acronym recall decrease. We therefore use $B=2$ as a practical trade-off between terminology recall and over-biasing risk.

\subsection{Shuffled-Memory Control}
\label{app:shuffled-control}

The shuffled-memory control tests whether terminology gains require matching the correct paper memory to the current speech stream. Table~\ref{tab:app_shuffled} reports the shuffled-memory diagnostic used in Section~\ref{sec:grounding-validation}.

\begin{table}[t]
\centering
\footnotesize
\setlength{\tabcolsep}{4pt}
\begin{tabular}{@{}lcc@{}}
\toprule
Condition & Anchor $\Delta$NER & Medium-evid. $\Delta$NER \\
\midrule
Correct memory & +0.266 & +58.2 \\
Shuffled memory & +0.012 & +7.5 \\
\bottomrule
\end{tabular}
\caption{
Shuffled-memory control on En$\rightarrow$De. $\Delta$NER is reported against the baseline. The large drop under shuffled memory indicates that the gains depend on matching the correct paper memory to the current speech stream.
}
\label{tab:app_shuffled}
\end{table}

\subsection{Reference-Supported Terminology Matching}
\label{app:precision-f1}

We further compute a reference-supported terminology matching diagnostic using the same normalization protocol as NER$_{\text{clean}}$. Gold terms are paper terms preserved in the reference, and predicted terms are paper terms realized in the system output. Because this diagnostic depends on reference-preserved surface forms, it is conservative and should be interpreted as a complementary surface-form analysis rather than as a direct measure of harmful over-generation. We do not use it as a direct over-generation measure because a system output can realize a valid paper term even when the single reference does not preserve the same surface form.

Under this protocol, EGTA-RG improves micro-F1 from 0.625 to 0.839 on En$\rightarrow$Zh and from 0.575 to 0.753 on En$\rightarrow$De. This suggests that EGTA improves reference-supported terminology coverage under a strict matching criterion.

We therefore use this diagnostic as a complement to NER$_{\text{clean}}$, acronym recall, activation audits, shuffled-memory controls, and over-bias stress tests, rather than as the primary ranking metric.

\subsection{SeamlessM4T Quality-Cost Decomposition}
\label{app:seamless-cost}

Section~\ref{sec:additional-analyses} reports terminology-fidelity transfer to SeamlessM4T-v2-large. Full quality metrics show small corpus-level changes: BLEU/chrF/XCOMET change by $-0.54/-0.33/-0.0028$ on En$\rightarrow$Zh and $-0.36/-0.89/-0.0084$ on En$\rightarrow$De. The cost is largely memory-agnostic: shuffled memory causes a similar quality change while removing most terminology gain. On segments whose references contain paper terms, EGTA-G improves chrF by +1.12 on En$\rightarrow$Zh and +1.02 on En$\rightarrow$De, with En$\rightarrow$De XCOMET also increasing by +0.0071.
\end{document}